\documentclass[10pt,british,runningheads]{llncs}
\usepackage[T1]{fontenc}
\usepackage[latin9]{inputenc}
\usepackage{babel}
\usepackage{array}
\usepackage{rotating}
\usepackage{wrapfig}
\usepackage{textcomp}
\usepackage{multirow}
\usepackage{amssymb}
\usepackage{graphicx}
\usepackage[unicode=true,pdfusetitle,
 bookmarks=true,bookmarksnumbered=false,bookmarksopen=false,
 breaklinks=false,pdfborder={0 0 1},backref=false,colorlinks=false]
 {hyperref}

\makeatletter

\providecommand{\tabularnewline}{\\}

\newenvironment{lyxlist}[1]
{\begin{list}{}
{\settowidth{\labelwidth}{#1}
 \setlength{\leftmargin}{\labelwidth}
 \addtolength{\leftmargin}{\labelsep}
 }}
{\end{list}}

\usepackage{graphicx}
\usepackage[table]{xcolor}

\usepackage{amsmath,amssymb} 

\usepackage[width=122mm,left=12mm,paperwidth=146mm,height=193mm,top=12mm,paperheight=217mm]{geometry}
\usepackage{rotating}

\usepackage{makecell, pict2e}

\usepackage{capt-of}

\definecolor{lightgray}{gray}{0.9}

\@ifundefined{showcaptionsetup}{}{%
 \PassOptionsToPackage{caption=false}{subfig}}
\usepackage{subfig}
\makeatother

\begin{document}
\pagestyle{headings} 
\mainmatter \def\ECCV14SubNumber{1}  

\title{Ten Years of Pedestrian Detection,\\
What Have We Learned?}

\titlerunning{Ten Years of Pedestrian Detection,What Have We Learned?}

\author{\hfill{}Rodrigo Benenson\hfill{}Mohamed Omran\hfill{}Jan Hosang\hfill{}Bernt
Schiele\hfill{}}

\authorrunning{Rodrigo Benenson, Mohamed Omran, Jan Hosang, and Bernt Schiele}

\institute{%
\begin{tabular}{c}
Max Planck Institute for Informatics\tabularnewline
Saarbrücken, Germany\tabularnewline
\texttt{\small{}firstname.lastname@mpi-inf.mpg.de}\tabularnewline
\end{tabular}}
\maketitle
\begin{abstract}

Paper-by-paper results make it easy to miss the forest for the trees.We
analyse the remarkable progress of the last decade by discussing the
main ideas explored in the 40+ detectors currently present in the
Caltech pedestrian detection benchmark. We observe that there exist
three families of approaches, all currently reaching similar detection
quality. Based on our analysis, we study the complementarity of the
most promising ideas by combining multiple published strategies. This
new decision forest detector achieves the current best known performance
on the challenging Caltech-USA dataset.\vspace{-1em}

\end{abstract}

\section{\label{sec:Introduction}Introduction}

\begin{wrapfigure}[10]{r}{0.6\columnwidth}%
\begin{centering}
\vspace{-5em}
\includegraphics[width=0.6\textwidth]{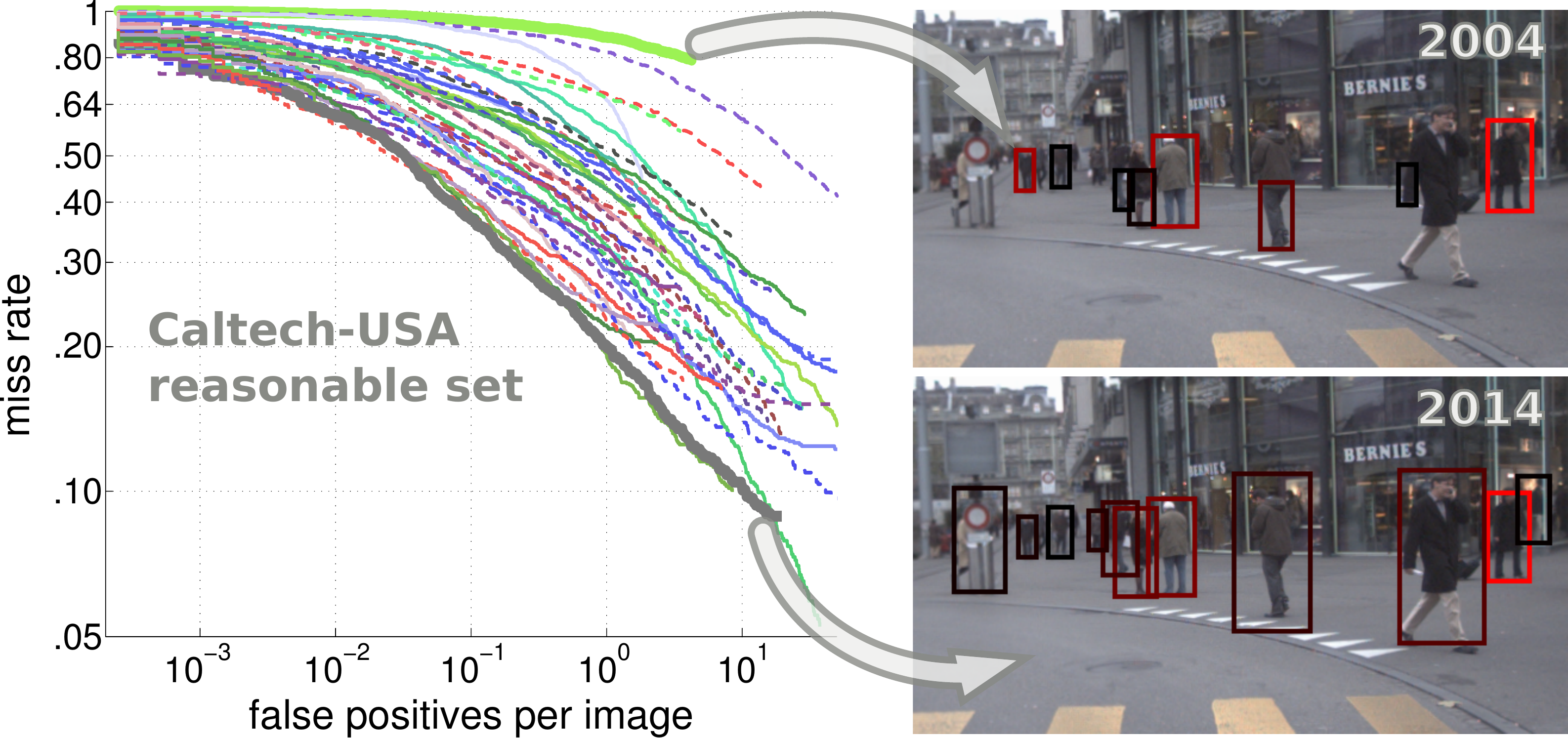}
\par\end{centering}

\begin{centering}
\vspace{-0.6em}

\par\end{centering}

\protect\caption{\label{fig:Sexy-figure}The last decade has shown tremendous progress
on pedestrian detection. What have we learned out of the $40+$ proposed
methods? }
\vspace{-2em}
\end{wrapfigure}%
Pedestrian detection is a canonical instance of object detection.
Because of its direct applications in car safety, surveillance, and
robotics, it has attracted much attention in the last years. Importantly,
it is a well defined problem with established benchmarks and evaluation
metrics. As such, it has served as a playground to explore different
ideas for object detection. The main paradigms for object detection
``Viola\&Jones variants'', HOG+SVM rigid templates, deformable part
detectors (DPM), and convolutional neural networks (ConvNets) have
all been explored for this task.

The aim of this paper is to review progress over the last decade of
pedestrian detection ($40+$ methods), identify the main ideas explored,
and try to quantify which ideas had the most impact on final detection
quality. In the next sections we review existing datasets (section
\ref{sec:Datasets}), provide a discussion of the different approaches
(section \ref{sec:Main-approaches}), and experiments reproducing/quantifying
the recent years' progress (section \ref{sec:Experiments}, presenting
experiments over $\sim20$ newly trained detector models). Although
we do not aim to introduce a novel technique, by putting together
existing methods we report the best known detection results on the
challenging Caltech-USA dataset.
\begin{figure}[t]
\begin{centering}
\vspace{-0.5em}
\hspace*{\fill}\subfloat[\label{fig:INRIA-detections}INRIA test set]{\begin{centering}
\includegraphics[width=0.325\textwidth]{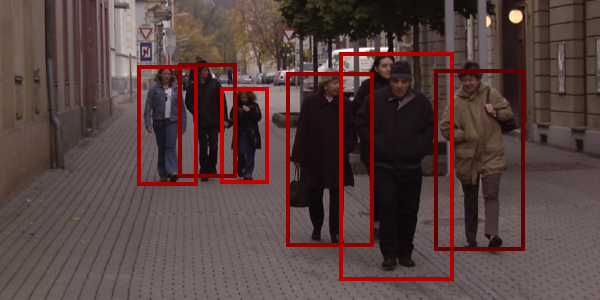}
\par\end{centering}

}\hspace*{\fill}\subfloat[\label{fig:Caltech-detections}Caltech-USA test set]{\begin{centering}
\includegraphics[width=0.325\textwidth]{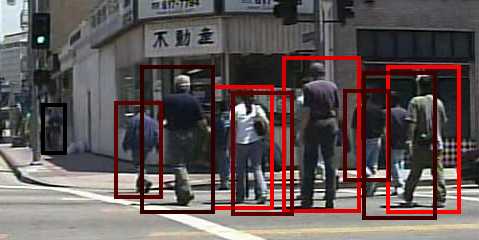}
\par\end{centering}

}\hspace*{\fill}\subfloat[\label{fig:KITTI-detections}KITTI test set]{\begin{centering}
\includegraphics[width=0.325\textwidth]{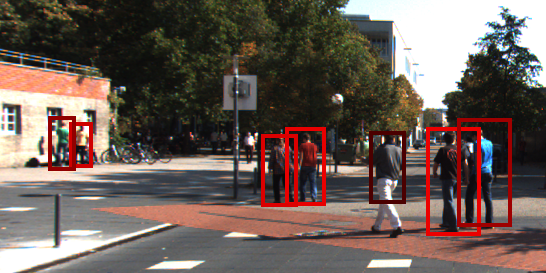}
\par\end{centering}

}\hspace*{\fill}
\par\end{centering}

\begin{centering}
\vspace{-1em}
\protect\caption{\label{fig:example-datasets}Example detections of a top performing
method (\texttt{SquaresChnFtrs}).}

\par\end{centering}

\centering{}\vspace{-1em}
\end{figure}

\section{\label{sec:Datasets}Datasets\vspace{-0.5em}
}

Multiple public pedestrian datasets have been collected over the
years; INRIA \cite{Dalal2005Cvpr}, ETH~\cite{Ess2008Cvpr}, TUD-Brussels~\cite{Wojek2009Cvpr},
Daimler \cite{Enzweiler2009PAMI} (Daimler stereo \cite{Keller2009Dagm}),
Caltech-USA~\cite{Dollar2009Cvpr}, and KITTI \cite{Geiger2012CVPR}
are the most commonly used ones. They all have different characteristics,
weaknesses, and strengths. 

INRIA is amongst the oldest and as such has comparatively few images.
It benefits however from high quality annotations of pedestrians in
diverse settings (city, beach, mountains, etc.), which is why it
is commonly selected for training (see also \S\ref{sub:Generalization-across-datasets}).
 ETH and TUD-Brussels are mid-sized video datasets. Daimler is not
considered by all methods because it lacks colour channels. Daimler
stereo, ETH, and KITTI provide stereo information.  All datasets
but INRIA are obtained from video, and thus enable the use of optical
flow as an additional cue.

Today, Caltech-USA and KITTI are the predominant benchmarks for pedestrian
detection. Both are comparatively large and challenging. Caltech-USA
stands out for the large number of methods that have been evaluated
side-by-side. KITTI stands out because its test set is slightly more
diverse, but is not yet used as frequently. For a more detailed discussion
of the datasets please consult \cite{Dollar2011Pami,Geiger2012CVPR}.
INRIA, ETH (monocular), TUD-Brussels, Daimler (monocular), and Caltech-USA
are available under a unified evaluation toolbox; KITTI uses its own
separate one with unpublished test data. Both toolboxes maintain an
online ranking where published methods can be compared side by side.

In this paper we use primarily Caltech-USA for comparing methods,
INRIA and KITTI secondarily. See figure \ref{fig:example-datasets}
for example images.  Caltech-USA and INRIA results are measured in
log-average miss-rate (MR, lower is better), while KITTI uses area
under the precision-recall curve (AUC, higher is better).
\begin{table}
\begin{centering}
\vspace{-1.5em}

\par\end{centering}

\begin{centering}
\rowcolors{2}{white}{lightgray}%
\begin{tabular}{rc|cccccccc|c|l}
Method & MR & \begin{turn}{70}
{\scriptsize{}Family}
\end{turn} & \begin{turn}{70}
{\scriptsize{}Features}
\end{turn} & \begin{turn}{70}
{\scriptsize{}Classifier}
\end{turn} & \begin{turn}{70}
{\scriptsize{}Context}
\end{turn} & \begin{turn}{70}
{\scriptsize{}Deep}
\end{turn} & \begin{turn}{70}
{\scriptsize{}Parts}
\end{turn} & \begin{turn}{70}
{\scriptsize{}M-Scales}
\end{turn} & \begin{turn}{70}
{\scriptsize{}More data}
\end{turn} & \begin{turn}{70}
{\scriptsize{}Feat. type}
\end{turn} & \begin{turn}{70}
{\scriptsize{}Training}
\end{turn}\tabularnewline
\hline 
\hline 
\texttt{VJ\hspace{0.7em}\cite{Viola2004Ijvc}} & $94.73\%$ & DF & $\checkmark$ & $\checkmark$ &  &  &  &  &  & {\scriptsize{}Haar} & ~I\tabularnewline
\texttt{Shapelet\,\cite{Sabzmeydani2007Cvpr}} & $91.37\%$ & - & $\checkmark$ &  &  &  &  &  &  & {\scriptsize{}Gradients} & ~I\tabularnewline
\texttt{PoseInv\,\cite{Lin2008Eccv}} & $86.32\%$ & - &  &  &  &  & $\checkmark$ &  &  & {\scriptsize{}HOG} & ~I+\tabularnewline
\texttt{LatSvm-V1\,\cite{Felzenszwalb2008CVPR}} & $79.78\%$ & DPM &  &  &  &  & $\checkmark$ &  &  & {\scriptsize{}HOG} & ~P\tabularnewline
\texttt{ConvNet\,\cite{Sermanet2013Cvpr}} & $77.20\%$ & DN &  &  &  & $\checkmark$ &  &  &  & {\scriptsize{}Pixels} & ~I\tabularnewline
\texttt{FtrMine\,\cite{Dollar2007Cvpr}} & $74.42\%$ & DF & $\checkmark$ &  &  &  &  &  &  & {\scriptsize{}HOG+Color } & ~I\tabularnewline
\texttt{HikSvm\,\cite{Maji2008Cvpr}} & $73.39\%$ & - &  & $\checkmark$ &  &  &  &  &  & {\scriptsize{}HOG} & ~I\tabularnewline
\texttt{HOG\hspace{0.7em}\cite{Dalal2005Cvpr}} & $68.46\%$ & - & $\checkmark$ & $\checkmark$ &  &  &  &  &  & {\scriptsize{}HOG} & ~I\tabularnewline
\texttt{MultiFtr\,\cite{Wojek2008DagmMultiFtrs}} & $68.26\%$ & DF & $\checkmark$ & $\checkmark$ &  &  &  &  &  & {\scriptsize{}HOG+Haar } & ~I\tabularnewline
\texttt{HogLbp\,\cite{Wang2009Iccv}} & $67.77\%$ & - & $\checkmark$ &  &  &  &  &  &  & {\scriptsize{}HOG+LBP} & ~I\tabularnewline
\texttt{AFS+Geo\,\cite{Levi2013Cvpr}} & $66.76\%$ & - &  &  & $\checkmark$ &  &  &  &  & {\scriptsize{}Custom } & ~I\tabularnewline
\texttt{AFS\,\cite{Levi2013Cvpr}} & $65.38\%$ & - &  &  &  &  &  &  &  & {\scriptsize{}Custom } & ~I\tabularnewline
\texttt{LatSvm-V2\,\cite{Felzenszwalb2010Pami}} & $63.26\%$ & DPM &  & $\checkmark$ &  &  & $\checkmark$ &  &  & {\scriptsize{}HOG} & ~I\tabularnewline
\texttt{Pls\,\cite{Schwartz2009Iccv}} & $62.10\%$ & - & $\checkmark$ & $\checkmark$ &  &  &  &  &  & {\scriptsize{}Custom} & ~I\tabularnewline
\texttt{MLS\,\cite{Nam2011IccvWorkshop}} & $61.03\%$ & DF & $\checkmark$ &  &  &  &  &  &  & {\scriptsize{}HOG} & ~I\tabularnewline
\texttt{MultiFtr+CSS\,\cite{Walk2010Cvpr}} & $60.89\%$ & DF & $\checkmark$ &  &  &  &  &  &  & {\scriptsize{}Many} & ~T\tabularnewline
\texttt{FeatSynth\,\cite{BarHillel2010Eccv}} & $60.16\%$ & - & $\checkmark$ & $\checkmark$ &  &  &  &  &  & {\scriptsize{}Custom } & ~I\tabularnewline
\texttt{pAUCBoost\,\cite{Paisitkriangkrai2013Iccv}} & $59.66\%$ & DF & $\checkmark$ & $\checkmark$ &  &  &  &  &  & {\scriptsize{}HOG+COV} & ~I\tabularnewline
\texttt{FPDW\,\cite{Dollar2010Bmvc}} & $57.40\%$ & DF &  &  &  &  &  &  &  & {\scriptsize{}HOG+LUV} & ~I\tabularnewline
\texttt{ChnFtrs\,\cite{Dollar2009Bmvc}} & $56.34\%$ & DF & $\checkmark$ & $\checkmark$ &  &  &  &  &  & {\scriptsize{}HOG+LUV} & ~I\tabularnewline
\texttt{CrossTalk\,\cite{Dollar2012Eccv}} & $53.88\%$ & DF &  &  & $\checkmark$ &  &  &  &  & {\scriptsize{}HOG+LUV} & ~I\tabularnewline
\texttt{DBN\textminus Isol\,\cite{Ouyang2012Cvpr}} & $53.14\%$ & DN &  &  &  &  & $\checkmark$ &  &  & {\scriptsize{}HOG} & ~I\tabularnewline
\texttt{ACF\,\cite{Dollar2014Pami}} & $51.36\%$ & DF & $\checkmark$ &  &  &  &  &  &  & {\scriptsize{}HOG+LUV} & ~I\tabularnewline
\texttt{RandForest\,\cite{Marin2013Iccv}} & $51.17\%$ & DF &  & $\checkmark$ &  &  &  &  &  & {\scriptsize{}HOG+LBP} & ~I\&C\tabularnewline
\texttt{MultiFtr+Motion\,\cite{Walk2010Cvpr}} & $50.88\%$ & DF & $\checkmark$ &  &  &  &  &  & $\checkmark$ & {\scriptsize{}Many+Flow } & ~T\tabularnewline
\texttt{\emph{\footnotesize{}SquaresChnFtrs}}\texttt{\,\cite{Benenson2013Cvpr}} & $50.17\%$ & DF & $\checkmark$ &  &  &  &  &  &  & {\scriptsize{}HOG+LUV} & ~I\tabularnewline
\texttt{Franken\,\cite{Mathias2013Iccv}} & $48.68\%$ & DF &  & $\checkmark$ &  &  &  &  &  & {\scriptsize{}HOG+LUV} & ~I\tabularnewline
\texttt{MultiResC\,\cite{Park2010Eccv}} & $48.45\%$ & DPM &  &  & $\checkmark$ &  & $\checkmark$ & $\checkmark$ &  & {\scriptsize{}HOG} & ~C\tabularnewline
\texttt{Roerei\,\cite{Benenson2013Cvpr}} & $48.35\%$ & DF & $\checkmark$ &  &  &  &  & $\checkmark$ &  & {\scriptsize{}HOG+LUV} & ~I\tabularnewline
\texttt{DBN\textminus Mut\,\cite{Ouyang2013CvprDbnMut}} & $48.22\%$ & DN &  &  & $\checkmark$ &  & $\checkmark$ &  &  & {\scriptsize{}HOG} & ~C\tabularnewline
\texttt{MF+Motion+2Ped\,\cite{Ouyang2013Cvpr}} & $46.44\%$ & DF &  &  & $\checkmark$ &  &  &  & $\checkmark$ & {\scriptsize{}Many+Flow } & ~I+\tabularnewline
\texttt{MOCO\,\cite{Chen2013Cvpr}} & $45.53\%$ & - & $\checkmark$ &  & $\checkmark$ &  &  &  &  & {\scriptsize{}HOG+LBP} & ~C\tabularnewline
\texttt{MultiSDP\,\cite{Zeng2013Iccv}} & $45.39\%$ & DN & $\checkmark$ &  & $\checkmark$ & $\checkmark$ &  &  &  & {\scriptsize{} HOG+CSS} & ~C\tabularnewline
\texttt{ACF-Caltech\,\cite{Dollar2014Pami}} & $44.22\%$ & DF & $\checkmark$ &  &  &  &  &  &  & {\scriptsize{}HOG+LUV} & ~C\tabularnewline
\texttt{MultiResC+2Ped\,\cite{Ouyang2013Cvpr}} & $43.42\%$ & DPM &  &  & $\checkmark$ &  & $\checkmark$ & $\checkmark$ &  & {\scriptsize{}HOG} & ~C+\tabularnewline
\texttt{WordChannels\,\cite{Costea2014CVPR}} & $42.30\%$ & DF & $\checkmark$ &  &  &  &  &  &  & {\scriptsize{}Many} & ~C\tabularnewline
\texttt{MT-DPM\,\cite{Yan2013Cvpr}} & $40.54\%$ & DPM &  &  &  &  & $\checkmark$ & $\checkmark$ &  & {\scriptsize{}HOG} & ~C\tabularnewline
\texttt{JointDeep\,\cite{Ouyang2013Iccv}} & $39.32\%$ & DN &  &  & $\checkmark$ &  &  &  &  & {\scriptsize{}Color+Gradient } & ~C\tabularnewline
\texttt{SDN\,\cite{Luo2014Cvpr}} & $37.87\%$ & DN &  &  &  & $\checkmark$ & $\checkmark$ &  &  & {\scriptsize{}Pixels} & ~C\tabularnewline
\texttt{MT-DPM+Context\,\cite{Yan2013Cvpr}} & $37.64\%$ & DPM &  &  & $\checkmark$ &  & $\checkmark$ & $\checkmark$ &  & {\scriptsize{}HOG} & ~C+\tabularnewline
\texttt{ACF+SDt\,\cite{Park2013Cvpr}} & $37.34\%$ & DF & $\checkmark$ &  &  &  &  &  & $\checkmark$ & \texttt{\scriptsize{}ACF}{\scriptsize{}+Flow} & ~C+\tabularnewline
\texttt{\emph{\footnotesize{}SquaresChnFtrs}}\texttt{\,\cite{Benenson2013Cvpr}} & $34.81\%$ & DF & $\checkmark$ &  &  &  &  &  &  & {\scriptsize{}HOG+LUV} & ~C\tabularnewline
\texttt{InformedHaar\,\cite{Zhang2014CvprInformedHaar}} & $34.60\%$ & DF & $\checkmark$ &  &  &  &  &  &  & {\scriptsize{}HOG+LUV} & ~C\tabularnewline
\texttt{\emph{Katamari-v1}} & $22.49\%$ & DF & $\checkmark$ &  & $\checkmark$ &  &  &  & $\checkmark$ & {\scriptsize{}HOG+Flow} & ~C+\tabularnewline
\end{tabular}
\par\end{centering}

\vspace{0.5em}

\protect\caption{\label{tab:caltech-methods}Listing of methods considered on Caltech-USA,
sorted by log-average miss-rate (lower is better). Consult sections
\ref{sub:Training-data} to \ref{sub:Better-features} for details
of each column. See also matching figure \ref{fig:caltech-usa-training-methods}.
\texttt{}``HOG'' indicates \texttt{HOG}-like \cite{Dalal2005Cvpr}.
Ticks indicate salient aspects of each method.}
\vspace{-0em}
\end{table}
 \vspace{-1.5em}

\subsubsection{\label{sub:Value-of-benchmarks}Value of benchmarks}

Individual papers usually only show a narrow view over the state of
the art on a dataset.  Having an official benchmark that collects
detections from all methods greatly eases the author's effort to put
their curve into context, and provides reviewers easy access to the
state of the art results. The collection of results enable retrospective
analyses such as the one presented in the next section.\vspace{-1em}

\section{\label{sec:Main-approaches}Main approaches to improve pedestrian
detection}

\vspace{-0.5em}

\begin{wrapfigure}{r}{0.63\columnwidth}%
\centering{}\vspace{-4em}
\hspace*{-1em}\includegraphics[width=0.63\columnwidth]{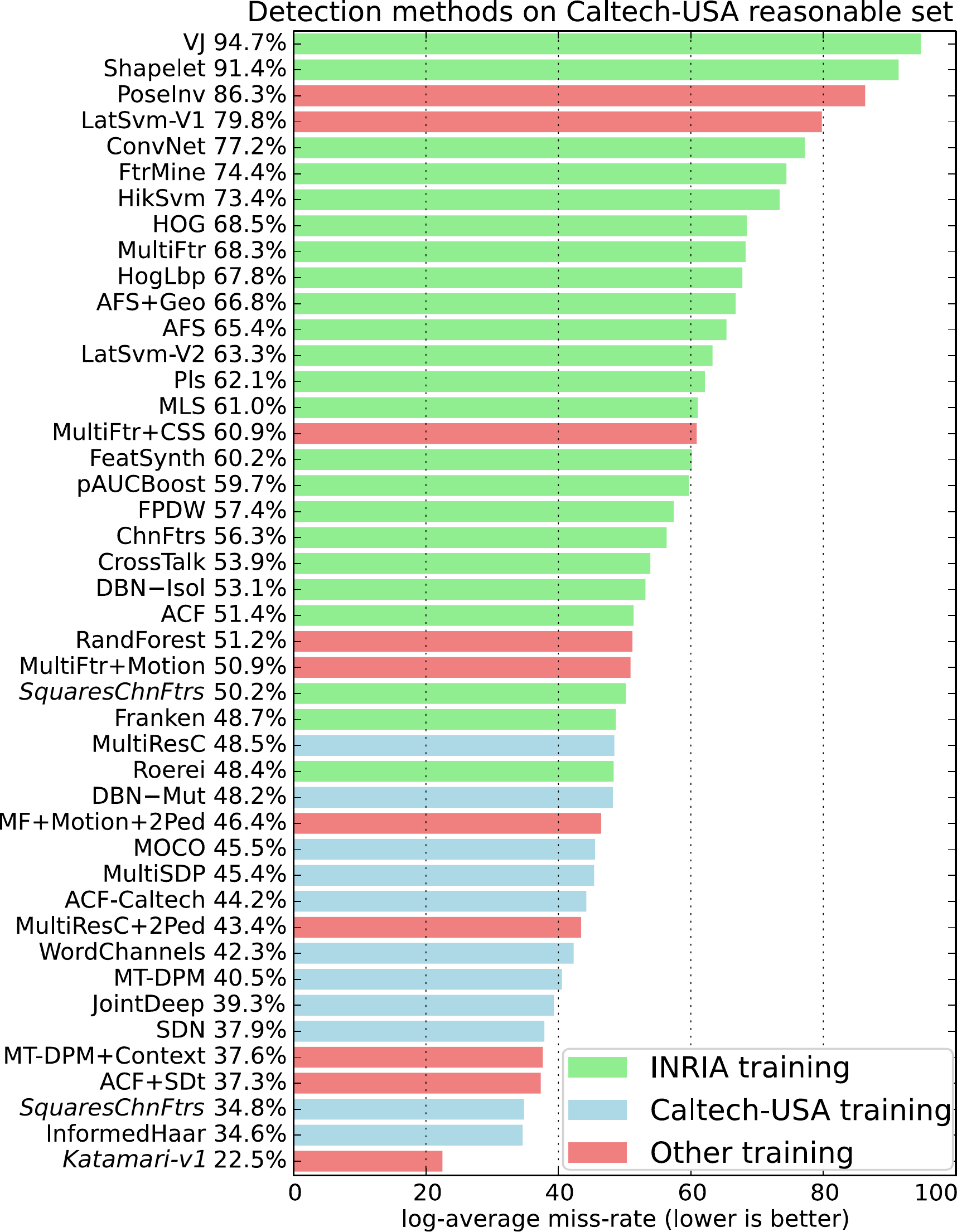}\vspace{-0.5em}
\protect\caption{\label{fig:caltech-usa-training-methods}Caltech-USA detection results.}
\vspace{-2em}
\end{wrapfigure}

Figure \ref{fig:caltech-usa-training-methods} and table \ref{tab:caltech-methods}
together provide a quantitative and qualitative overview over $40+$
methods whose results are published on the Caltech pedestrian detection
benchmark (July 2014). Methods marked in italic are our newly trained
models (described in section \ref{sec:Experiments}). We refer to
all methods using their Caltech benchmark shorthand. Instead of discussing
the methods' individual particularities, we identify the key aspects
that distinguish each method (ticks of table \ref{tab:caltech-methods})
and group them accordingly. We discuss these aspects in the next subsections.

\subsubsection{\label{sub:Brief-chronology}Brief chronology}

In 2003, Viola and Jones applied their \texttt{VJ} detector \cite{Viola2003Cvpr}
to the task of pedestrian detection. In 2005 Dalal and Triggs introduced
the landmark \texttt{HOG} \cite{Dalal2005Cvpr} detector, which later
served in 2008 as a building block for the now classic deformable
part model DPM (named \texttt{LatSvm} here) by Felzen\-swalb et al.~\cite{Felzenszwalb2008CVPR}.
In 2009 the Caltech pedestrian detection benchmark was introduced,
comparing seven pedestrian detectors \cite{Dollar2009Cvpr}. At this
point in time, the evaluation metrics changed from per-window (FPPW)
to per-image (FPPI), once the flaws of the per-window evaluation were
identified \cite{Dollar2011Pami}. Under this new evaluation metric
some of the early detectors turned out to under-perform.\\
About one third of the methods considered here were published during
2013, reflecting a renewed interest on the problem. Similarly, half
of the KITTI results for pedestrian detection were submitted in 2014.\vspace{-1em}

\subsection{\label{sub:Training-data}Training data}

Figure \ref{fig:caltech-usa-training-methods} shows that differences
in detection performance are, not surprisingly, dominated by the choice
of training data. Methods trained on Caltech-USA systematically
perform better than methods that generalise from INRIA. Table \ref{tab:caltech-methods}
gives additional details on the training data used%
\footnote{\texttt{ }``Training'' data column: I$\rightarrow$INRIA, C$\rightarrow$Caltech,
I+/C+ $\rightarrow$INRIA/Caltech and additional data, P$\rightarrow$Pascal,
T$\rightarrow$TUD-Motion, I\&C$\rightarrow$both INRIA and Caltech.%
}. High performing methods with ``other training'' use extended versions
of Caltech-USA. For instance \texttt{MultiResC+2Ped} uses Caltech-USA
plus an extended set of annotations over INRIA, \texttt{MT-DPM+Context}
uses an external training set for cars, and \texttt{ACF+SDt} employs
additional frames from the original Caltech-USA videos.\\

\subsection{\label{sub:Solution-families}Solution families}

Overall we notice that out of the $40+$ methods we can discern three
families: 1) DPM variants (\texttt{MultiResC} \cite{Park2010Eccv},
\texttt{MT-DPM} \cite{Yan2013Cvpr}, etc.), 2) Deep networks (\texttt{JointDeep}
\cite{Ouyang2013Iccv}, \texttt{ConvNet} \cite{Sermanet2013Cvpr},
etc.), and 3) Decision forests (\texttt{ChnFtrs}, \texttt{Roerei},
etc.). On table \ref{tab:caltech-methods} we identify these families
as \texttt{DPM}, \texttt{DN}, and \texttt{DF} respectively.

Based on raw numbers alone boosted decision trees (\texttt{DF}) seem
particularly suited for pedestrian detection, reaching top performance
on both the ``train on INRIA, test on Caltech'', and ``train on
Caltech, test on Caltech'' tasks. It is unclear however what gives
them an edge. The deep networks explored also show interesting properties
and fast progress in detection quality.\vspace{-1em}

\paragraph{Conclusion}

Overall, at present, DPM variants, deep networks, and (boosted) decision
forests all reach top performance in pedestrian detection (around
$37\,\%$ MR on Caltech-USA, see figure~\ref{fig:caltech-usa-training-methods}).

\subsection{\label{sub:Better-classifiers}Better classifiers}

Since the original proposal of \texttt{HOG+SVM} \cite{Dalal2005Cvpr},
linear and non-linear kernels have been considered. \texttt{HikSvm
}\cite{Maji2008Cvpr} considered fast approximations of non-linear
kernels. This method obtains improvements when using the flawed FPPW
evaluation metric (see section \ref{sec:Main-approaches}), but fails
to perform well under the proper evaluation (FPPI). In the work on
\texttt{MultiFtrs} \cite{Wojek2008DagmMultiFtrs}, it was argued that,
given enough features, Adaboost and linear SVM perform roughly the
same for pedestrian detection.

Recently, more and more components of the detector are optimized
jointly with the ``decision component'' (e.g. pooling regions in
\texttt{ChnFtrs}~\cite{Dollar2009Bmvc}, filters in \texttt{JointDeep}~\cite{Ouyang2013Iccv}).
As a result the distinction between features and classifiers is not
clear-cut anymore (see also sections \ref{sub:Deep-architectures}
and \ref{sub:Better-features}).

\vspace{-1em}

\paragraph{Conclusion}

There is no conclusive empirical evidence indicating that whether
non-linear kernels provide meaningful gains over linear kernels (for
pedestrian detection, when using non-trivial features). Similarly,
it is unclear whether one particular type of classifier (e.g. SVM
or decision forests) is better suited for pedestrian detection than
another.

\subsection{\label{sub:additional-data}Additional data}

The core problem of pedestrian detection focuses on individual monocular
colour image frames. Some methods explore leveraging additional information
at training and test time to improve detections. They consider stereo
images \cite{Keller2011Its}, optical flow (using previous frames,
e.g. \texttt{MultiFtr+Motion} \cite{Walk2010Cvpr} and \texttt{ACF+SDt}
\cite{Park2013Cvpr}), tracking \cite{Ess2009Pami}, or data from
other sensors (such as lidar \cite{Premebida2014Iros} or radar).

\begin{wrapfigure}{r}{0.55\columnwidth}%
\centering{}\vspace{-1em}
\hspace*{-0.5em}\includegraphics[width=0.6\columnwidth]{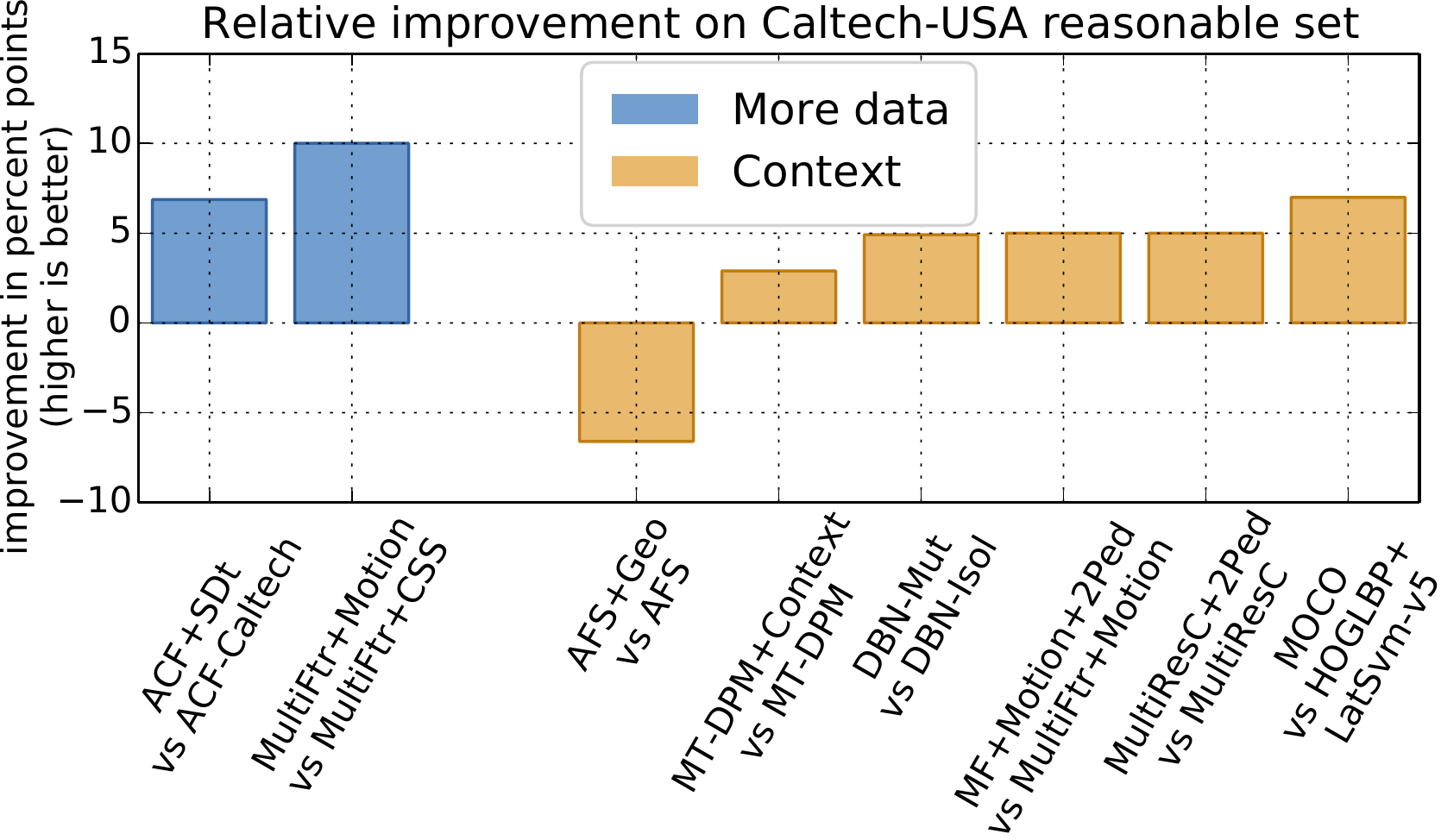}\vspace{-1em}
\protect\caption{\label{fig:caltech-usa-flow-context-deep}Caltech-USA detection improvements
for different method types. Improvement relative to each method's
relevant baseline (``method vs baseline'').}
\vspace{-4em}
\end{wrapfigure}

For monocular methods it is still unclear how much tracking can improve
per-frame detection itself. As seen in figure \ref{fig:caltech-usa-flow-context-deep}
exploiting optical flow provides a non-trivial improvement over the
baselines. Curiously, the current best results \texttt{(ACF-SDt}
\cite{Park2013Cvpr}) are obtained using coarse rather than high quality
flow.  In section \ref{sub:Complementarity-experiments} we  inspect
the complementarity of flow with other ingredients. Good success exploiting
flow and stereo on the Daimler dataset has been reported \cite{Enzweiler2011ImageProcessing},
but similar results have yet to be seen on newer datasets such as
KITTI.

\vspace{-1em}

\paragraph{Conclusion}

Using additional data provides meaningful improvements, albeit on
modern dataset stereo and flow cues have yet to be fully exploited.
As of now, methods based merely on single monocular image frames have
been able to keep up with the performance improvement introduced by
additional information.%

\subsection{\label{sub:Exploiting-context}Exploiting context}

Sliding window detectors score potential detection windows using the
content inside that window. Drawing on the context of the detection
window, i.e. image content surrounding the window, can improve detection
performance. Strategies for exploiting context include: ground plane
constraints (\texttt{Multi\-ResC} \cite{Park2010Eccv}, \texttt{Rand\-Forest}
\cite{Marin2013Iccv}), variants of auto-context \cite{Tu2010Pami}
(\texttt{MOCO} \cite{Chen2013Cvpr}), other category detectors (\texttt{MT-DPM\-+\-Context}
\cite{Yan2013Cvpr}), and person-to-person patterns (\texttt{DBN\textminus \-Mut}
\cite{Ouyang2013CvprDbnMut},\texttt{ +2Ped} \cite{Ouyang2013Cvpr},\texttt{
Joint\-Deep} \cite{Ouyang2013Iccv})\texttt{.}

Figure \ref{fig:caltech-usa-flow-context-deep} shows the performance
improvement for methods incorporating context.   Overall, we see
improvements of $3\sim7$ MR percent points. \texttt{(}The negative
impact of \texttt{AFS+Geo} is due to a change in evaluation, see section
\ref{sub:Brief-chronology}.) Interestingly, \texttt{+2Ped} \cite{Ouyang2013Cvpr}
obtains a consistent $2\sim5$ MR percent point improvement over existing
methods, even top performing ones (see section \ref{sub:Complementarity-experiments}).

\vspace{-1em}

\paragraph{Conclusion}

Context provides consistent improvements for pedestrian detection,
although the scale of improvement is lower compared to additional
test data (\S\ref{sub:additional-data}) and deep architectures (\S\ref{sub:Deep-architectures}).
The bulk of detection quality must come from other sources.

\subsection{\label{sub:Deformable-parts}Deformable parts}

The DPM detector \cite{Felzenszwalb2010Pami} was originally motivated
for pedestrian detection. It is an idea that has become very popular
and dozens of variants have been explored.

For pedestrian detection the results are competitive, but not salient
(\texttt{LatSvm }\cite{Yan2014Cvpr,Felzenszwalb2008CVPR}, \texttt{MultiResC}
\cite{Park2010Eccv}, \texttt{MT-DPM} \cite{Yan2013Cvpr}). More interesting
results have been obtained when modelling parts and their deformations
inside a deep architecture (e.g.\texttt{ DBN\textminus \-Mut} \cite{Ouyang2013CvprDbnMut},
\texttt{Joint\-Deep} \cite{Ouyang2013Iccv}). 

\texttt{DPM }and its variants are systematically outmatched by methods
using a single component and no parts (\texttt{Roe\-rei} \cite{Benenson2013Cvpr},
\texttt{Squares\-ChnFtrs }see section \ref{sub:Feature-experiments}\texttt{),
}casting doubt on the need for parts. Recent work has explored ways
to capture deformations entirely without parts \cite{Hariharan2014Cvpr,Pedersoli2014Cvpr}.

\vspace{-1em}

\paragraph{Conclusion}

For pedestrian detection there is still no clear evidence for the
necessity of components and parts, beyond the case of occlusion handling.

\subsection{\label{sub:Multi-scales}Multi-scale models}

Typically for detection, both high and low resolution candidate windows
are resampled to a common size before extracting features. It has
recently been noticed that training different models for different
resolutions systematically improve performance by $1\sim2$ MR percent
points \cite{Park2010Eccv,Benenson2013Cvpr,Yan2013Cvpr}, since the
detector has access to the full information available at each window
size. This technique does not impact computational cost at detection
time \cite{Benenson2012Cvpr}, although training time increases.

\vspace{-1em}

\paragraph{Conclusion}

Multi-scale models provide a simple and generic extension to existing
detectors. Despite consistent improvements, their contribution to
the final quality is rather minor.

\subsection{\label{sub:Deep-architectures}Deep architectures}

Large amounts of training data and increased computing power have
lead to recent successes of deep architectures (typically convolutional
neural networks) on diverse computer vision tasks (large scale classification
and detection \cite{Girshick2014ArxivRCNN,Sermanet2014Iclr}, semantic
labelling \cite{Pinheiro2014Jmlr}). These results have inspired the
application of deep architectures to the pedestrian task.

\texttt{ConvNet} \cite{Sermanet2013Cvpr} uses a mix of unsupervised
and supervised training to create a convolutional neural network trained
on INRIA. This method obtains fair results on INRIA, ETH, and TUD-Brussels,
however fails to generalise to the Caltech setup. This method learns
to extract features directly from raw pixel values.

Another line of work focuses on using deep architectures to jointly
model parts and occlusions (\texttt{DBN\textminus Isol} \cite{Ouyang2012Cvpr},
\texttt{DBN\textminus Mut} \cite{Ouyang2013CvprDbnMut}, \texttt{JointDeep}
\cite{Ouyang2013Iccv}, and \texttt{SDN} \cite{Luo2014Cvpr}). The
performance improvement such integration varies between $1.5$ to
$14$ MR percent points. Note that these works use edge and colour
features \cite{Ouyang2013Iccv,Ouyang2013CvprDbnMut,Ouyang2012Cvpr},
or initialise network weights to edge-sensitive filters, rather than
discovering features from raw pixel values as usually done in deep
architectures.\texttt{} No results have yet been reported using features
pre-trained on ImageNet \cite{Girshick2014ArxivRCNN,Azizpour2014arXiv}.

\vspace{-1em}

\paragraph{Conclusion}

Despite the common narrative there is still no clear evidence that
deep networks are good at learning features for pedestrian detection
(when using pedestrian detection training data). Most successful methods
use such architectures to model higher level aspects of parts, occlusions,
and context. The obtained results are on par with DPM and decision
forest approaches, making the advantage of using such involved architectures
yet unclear.

\subsection{\label{sub:Better-features}Better features}

The most popular approach (about $30\,\%$ of the considered methods)
for improving detection quality is to increase/diversify the features
computed over the input image. By having richer and higher dimensional
representations, the classification task becomes somewhat easier,
enabling improved results. A large set of feature types have been
explored: edge information \cite{Dalal2005Cvpr,Dollar2009Bmvc,Lim2013Cvpr,Luo2014Cvpr},
colour information \cite{Dollar2009Bmvc,Walk2010Cvpr}, texture information
\cite{Wang2009Iccv}, local shape information \cite{Costea2014CVPR},
covariance features \cite{Paisitkriangkrai2013Iccv}, amongst others.
More and more diverse features have been shown to systematically improve
performance. 

While various decision forest methods use $10$ feature channels (\texttt{ChnFtrs},
\texttt{ACF}, \texttt{Roerei}, \texttt{SquaresChnFtrs}, etc.), some
papers have considered up to an order of magnitude more channels \cite{Wojek2008DagmMultiFtrs,Lim2013Cvpr,Paisitkriangkrai2013Iccv,Marin2013Iccv,Costea2014CVPR}.
Despite the improvements by adding many channels, top performance
is still reached with only $10$ channels ($6$ gradient orientations,
$1$ gradient magnitude, and $3$ colour channels, we name these HOG+LUV);
see table \ref{tab:caltech-methods} and figure \ref{fig:caltech-usa-training-methods}.
In section \ref{sub:Feature-experiments} we study in more detail
different feature combinations.

From all what we see, from \texttt{VJ }($95\%$ MR) to \texttt{ChnFtrs}
($56.34\%$ MR, by adding HOG and LUV channels), to \texttt{SquaresChnFtrs-Inria}
($50.17\%$ MR, by exhaustive search over pooling sizes, see section
\ref{sec:Experiments}), improved features drive progress. Switching
training sets (section \ref{sub:Training-data}) enables \texttt{SquaresChnFtrs-Caltech}
to reach state of the art performance on the Caltech-USA dataset;
improving over significantly more sophisticated methods. \texttt{InformedHaar
}\cite{Zhang2014CvprInformedHaar} obtains top results by using a
set of Haar-like features manually designed for the pedestrian detection
task. In contrast \texttt{SquaresChnFtrs-Caltech} obtains similar
results without using such hand-crafted features and being data driven
instead.

Upcoming studies show that using more (and better features) yields
further improvements \cite{Paisitkriangkrai2014Eccv,Nam2014arXiv}.
It should be noted that better features for pedestrian detection have
not yet been obtained via deep learning approaches (see caveat on
ImageNet features in section \ref{sub:Deep-architectures}). \vspace{-1em}

\paragraph{Conclusion}

In the last decade improved features have been a constant driver for
detection quality improvement, and it seems that it will remain so
in the years to come. Most of this improvement has been obtained by
extensive trial and error. The next scientific step will be to develop
a more profound understanding of the what makes good features good,
and how to design even better ones%
\footnote{This question echoes with the current state of the art in deep learning,
too.%
}. 
\begin{figure}[t]
\centering{}%
\begin{minipage}[t]{0.49\columnwidth}%
\begin{center}
\includegraphics[width=1\textwidth]{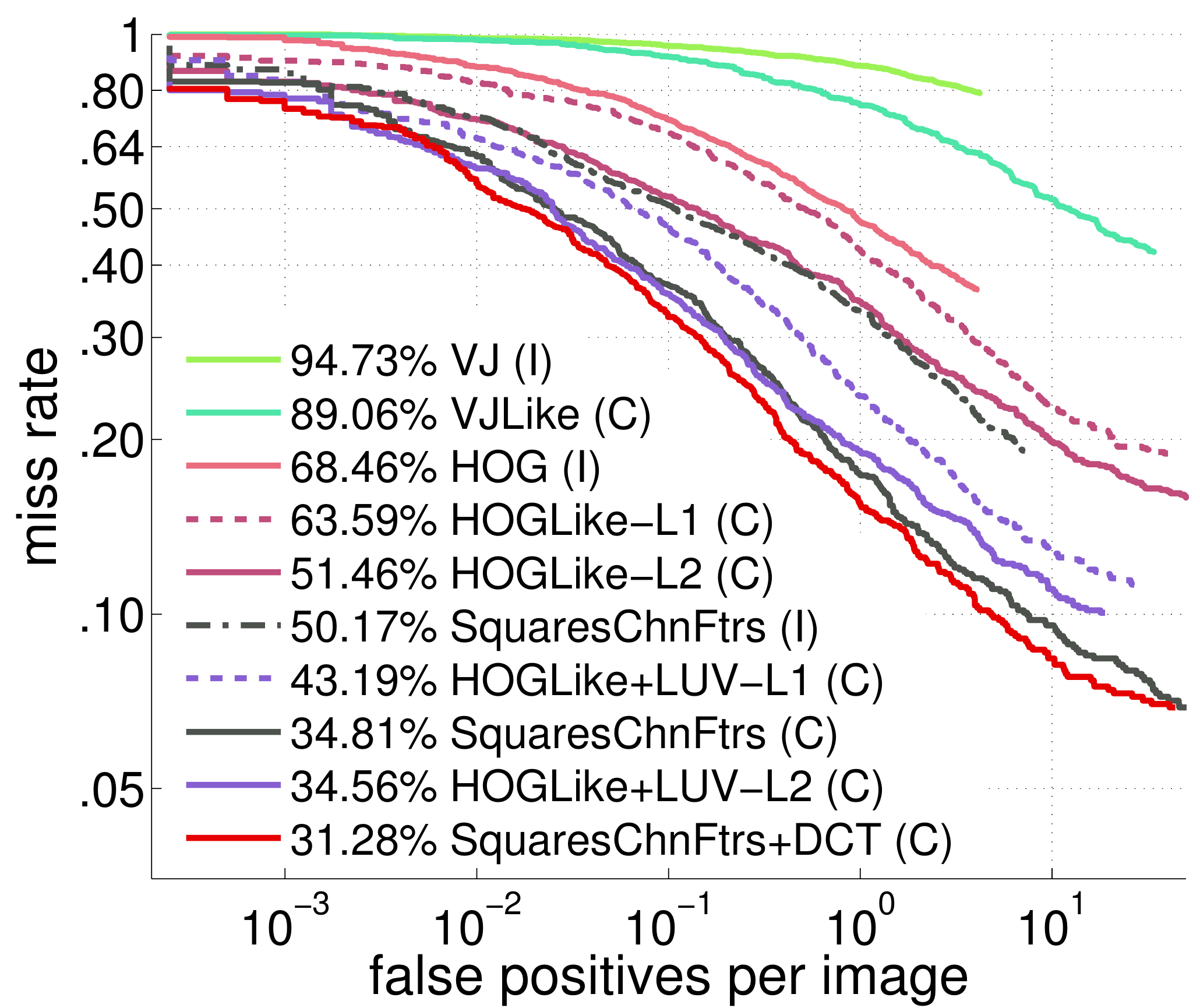}\vspace{-1em}

\par\end{center}

\begin{center}
\protect\caption{\label{fig:effect-of-features-Caltech}Effect of features on detection
performance. Caltech-USA reasonable test set.}

\par\end{center}%
\end{minipage}\hspace*{\fill}%
\begin{minipage}[t]{0.49\columnwidth}%
\begin{center}
\includegraphics[width=1\textwidth]{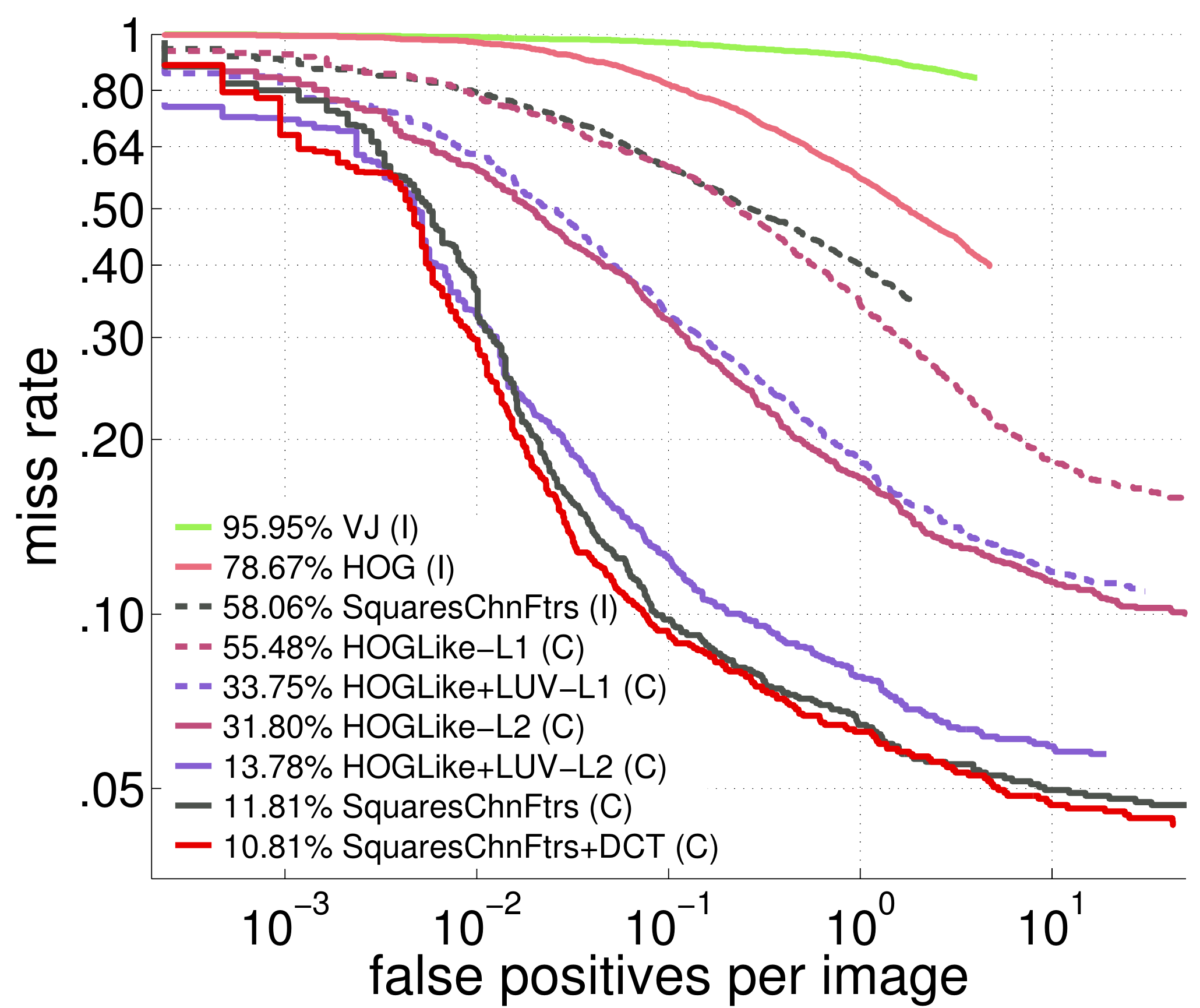}\protect\caption{\label{fig:caltech-train-set}Caltech-USA training set performance.
(I)/(C) indicates using INRIA/Caltech-USA training set.}

\par\end{center}%
\end{minipage}\vspace{-2em}
\end{figure}

\section{\label{sec:Experiments}Experiments}

Based on our analysis in the previous section, three aspects seem
to be the most promising in terms of impact on detection quality:
better features (\S\ref{sub:Better-features}), additional data (\S\ref{sub:additional-data}),
and context information (\S\ref{sub:Exploiting-context}). We thus
conduct experiments on the complementarity of these aspects.

Among the three solution families discussed (section \ref{sub:Solution-families}),
we choose the Integral Channels Features framework \cite{Dollar2009Bmvc}
(a decision forest) for conducting our experiments. Methods from this
family have shown good performance, train in minutes$\sim$hours,
and lend themselves to the analyses we aim.

In particular, we use the (open source) \texttt{SquaresChnFtrs} baseline
described in \cite{Benenson2013Cvpr}: $2048$ level-2 decision trees
($3$ threshold comparisons per tree) over \texttt{HOG+LUV} channels
($10$ channels), composing one $64\times128\ \mbox{pixels}$ template
learned via vanilla AdaBoost and few bootstrapping rounds of hard
negative mining.

\subsection{\label{sub:Feature-experiments}Reviewing the effect of features}

In this section, we evaluate the impact of increasing feature complexity.
We tune all methods on the INRIA test set, and demonstrate results
on the Caltech-USA test set (see figure \ref{fig:effect-of-features-Caltech}).
Results on INRIA as well as implementation details can be found in
the supplementary material.

The first series of experiments aims at mimicking landmark detection
techniques, such as \texttt{VJ}~\cite{Viola2003Cvpr}, \texttt{HOG}+linear
SVM~\cite{Dalal2005Cvpr}, and \texttt{ChnFtrs}~\cite{Dollar2009Bmvc}.
\texttt{VJLike} uses only the luminance colour channel, emulating
the Haar wavelet like features from the original \cite{Viola2003Cvpr}
using level 2 decision trees. \texttt{HOGLike-L1/L2 }use $8\times8\ \mbox{pixel}$
pooling regions, $1$ gradient magnitude and $6$ oriented gradient
channels, as well as level 1/2 decision trees. We also report results
when adding the LUV colour channels \texttt{HOGLike+LUV} ($10$ feature
channels total). \texttt{SquaresChnFtrs} is the baseline described
in the beginning of section \ref{sec:Experiments}, which is similar
to \texttt{HOGLike+LUV} to but with square pooling regions of any
size.

Inspired by \cite{Nam2014arXiv}, we also expand the $10$ HOG+LUV
channels into $40$ channels by convolving each channel with three
DCT (discrete cosine transform) basis functions (of $7\times7\ \mbox{pixels}$),
and storing the absolute value of the filter responses as additional
feature channels. We name this variant \texttt{SquaresChnFtrs+DCT}.

\paragraph{Conclusion}

Much of the progress since \texttt{VJ} can by explained by the use
of better features, based on oriented gradients and colour information.
Simple tweaks to these well known features (e.g. projection onto the
DCT basis) can still yield noticeable improvements.

\subsection{\label{sub:Complementarity-experiments}Complementarity of approaches}

\begin{figure}[t]
\begin{tabular}{ccc}
\begin{minipage}[t]{0.55\columnwidth}%
\begin{center}
\includegraphics[width=1\textwidth]{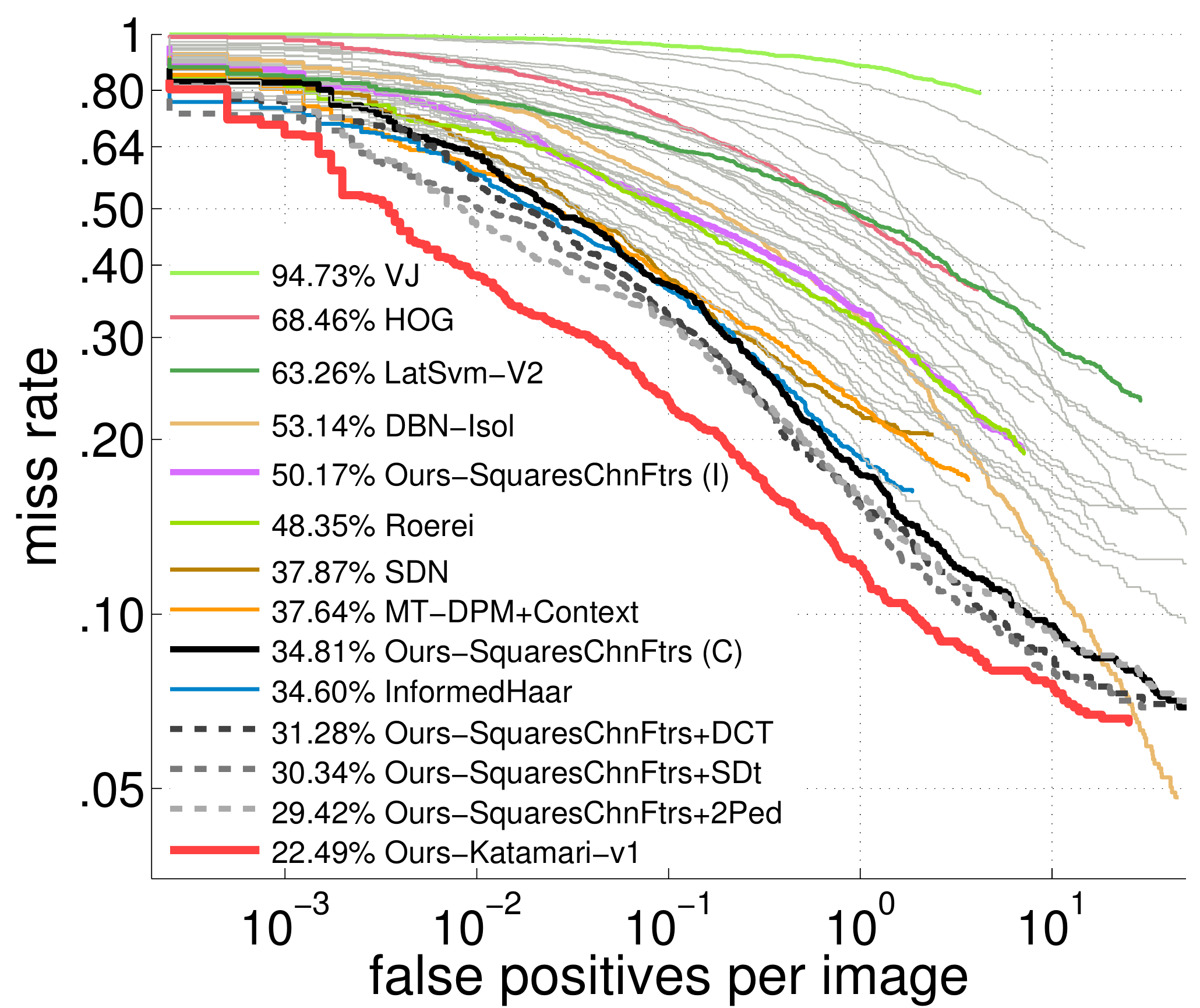}\vspace{-1em}

\par\end{center}

\caption{\label{fig:caltech-top-methods}Some of the top quality detection
methods for Caltech-USA. See section \ref{sub:Complementarity-experiments}.}
\end{minipage} & \enskip{} & %
\begin{minipage}[t][5em][s]{0.42\columnwidth}%
\includegraphics[width=1\textwidth]{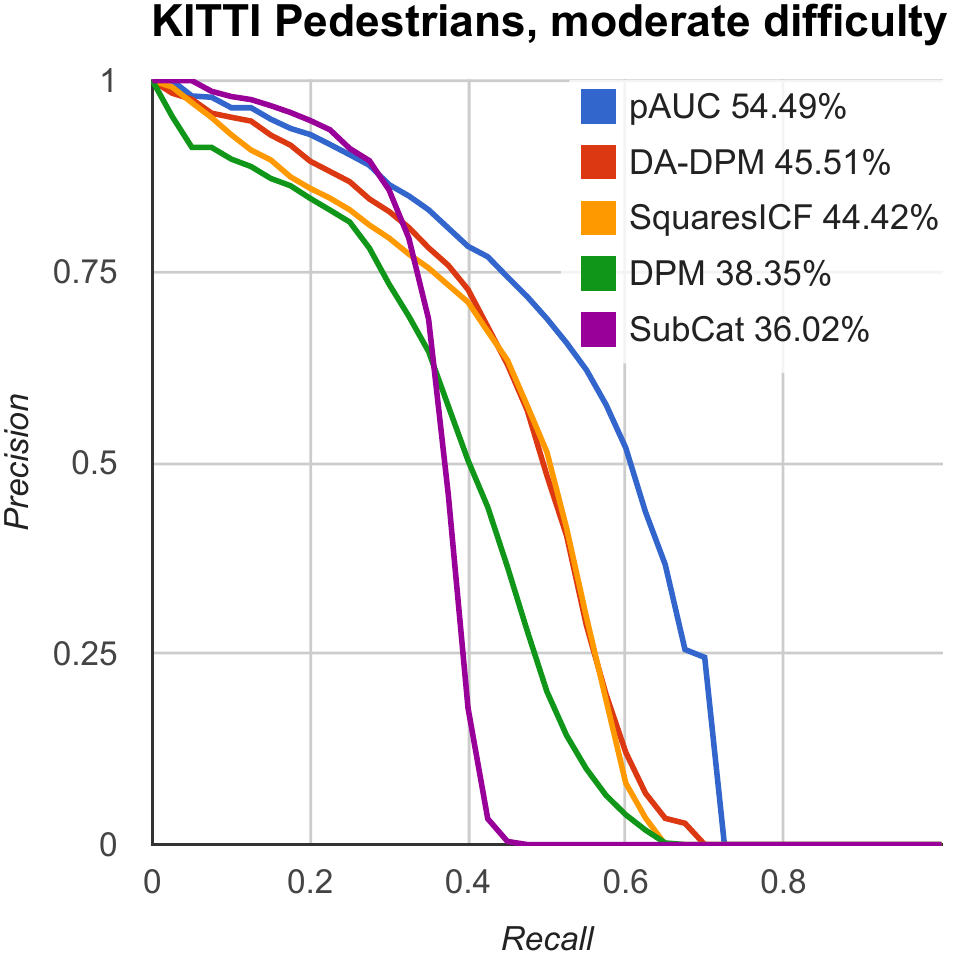}\vspace{-1em}

\caption{\label{fig:kitti-results}Pedestrian detection on the KITTI dataset.
}
\end{minipage}\tabularnewline
\end{tabular}
\end{figure}
After revisiting the effect of single frame features in section \ref{sub:Feature-experiments}
we now consider the complementary of better features (HOG+LUV+DCT),
additional data (via optical flow), and context (via person-to-person
interactions).

We encode the optical flow using the same SDt features from \texttt{ACF+SDt}
\cite{Park2010Eccv} (image difference between current frame T and
coarsely aligned T-4 and T-8). The context information is injected
using the \texttt{+2Ped} re-weighting strategy \cite{Ouyang2013Cvpr}
(the detection scores are combined with the scores of a ``2 person''
DPM detector). In all experiments both DCT and SDt features are pooled
over $8\times8$ regions (as in \cite{Park2010Eccv}), instead of
``all square sizes'' for the HOG+LUV features. 

The combination \texttt{SquaresChnFtrs+DCT+SDt+2Ped} is called \texttt{Katamari-v1}.
Unsurprisingly, \texttt{Katamari-v1} reaches the best known result
on the Caltech-USA dataset. In figure \ref{fig:caltech-top-methods}
we show it together with the best performing method for each training
set and solution family (see table \ref{tab:caltech-methods}). The
supplementary material contains results of all combinations between
the ingredients.

\vspace{-1em}

\paragraph{Conclusion}

Our experiments show that adding extra features, flow, and context
information are largely complementary ($12\,\%$ gain, instead of
$3+7+5\,\%$), even when starting from a strong detector.\\
It remains to be seen if future progress in detection quality will
be obtained by further insights of the ``core'' algorithm (thus
further diminishing the relative improvement of add-ons), or by extending
the diversity of techniques employed inside a system.

\subsection{\label{sub:capacity-experiments}How much model capacity is needed?}

The main task of detection is to generalise from training to test
set. Before we analyse the generalisation capability (section \ref{sub:Generalization-across-datasets}),
we consider a necessary condition for high quality detection: is the
learned model performing well on the training set?

In figure \ref{fig:caltech-train-set} we see the detection quality
of the models considered in section \ref{sub:Feature-experiments},
when evaluated over their training set. None of these methods performs
perfectly on the training set. In fact, the trend is very similar
to performance on the test set (see figure~\ref{fig:effect-of-features-Caltech})
and we do not observe yet symptoms of over-fitting. \vspace{-1em}

\paragraph{Conclusion}

Our results indicate that research on increasing the discriminative
power of detectors is likely to further improve detection quality.
More discriminative power can originate from more and better features
or more complex classifiers.

\subsection{\label{sub:Generalization-across-datasets}Generalisation\protect \\
across datasets}

\begin{wraptable}{r}{20.5em}%
\vspace{-7em}

\centering{}\protect\caption{\label{tab:Effect-of-training-set}Effect of training set over the
detection quality. Bold indicates second best training set for each
test set, except for ETH where bold indicates the best training set.}
\begin{tabular}{c|ccc}
\diaghead{\hskip7.2em}{Test\\set}{Training\\set} & INRIA & Caltech-USA & KITTI\tabularnewline
\hline 
\hline 
INRIA & $\mathit{17.42\,\%}$ & $60.50\,\%$ & $\mathbf{55.83\,\%}$\tabularnewline
Caltech-USA & $\mathbf{50.17\,\%}$ & $\mathit{34.81\,\%}$ & $61.19\,\%$\tabularnewline
KITTI & \textbf{$\mathbf{38.61\,\%}$} & $28.65\,\%$ & $\mathit{44.42\,\%}$\tabularnewline
ETH & $\mathbf{56.27\,\%}$ & $76.11\%$ & $61.19\,\%$\tabularnewline
\end{tabular}\vspace{-2em}
\end{wraptable}%
For real world application beyond a specific benchmark, the generalisation
capability of a model is key. In that sense re\-sults of models trained
on INRIA and tested on Caltech-USA are more relevant than the ones
trained (and tested) on Caltech-USA.

Table \ref{tab:Effect-of-training-set} shows the performance of \texttt{SquaresChnFtrs}
over Caltech-USA when using different training sets (MR for INRIA/Caltech/ETH,
AUC for KITTI{\small{})}. These experiments indicate that training
on Caltech or KITTI provides little generalisation capability towards
INRIA, while the converse is not true. Surprisingly, despite the
visual similarity between KITTI and Caltech, INRIA is the second best
training set choice for KITTI and Caltech. This shows that Caltech-USA
pedestrians are of ``their own kind'', and that the INRIA dataset
is effective due to its diversity. In other words few diverse pedestrians
(INRIA) is better than many similar ones (Caltech/KITTI). 

The good news is that the best methods seem to perform well both across
datasets and when trained on the respective training data. Figure
\ref{fig:kitti-results} shows methods trained and tested on KITTI,
we see that \texttt{SquaresChnFtrs} (named \texttt{SquaresICF} in
KITTI) is better than vanilla DPM and on par with the best known DPM
variant. The currently best method on KITTI, \texttt{pAUC} \cite{Paisitkriangkrai2014Eccv},
is a variant of \texttt{ChnFtrs} using $250$ feature channels (see
the KITTI website for details on the methods). These two observations
are consistent with our discussions in sections \ref{sub:Better-features}
and \ref{sub:Feature-experiments}.

\vspace{-1em}

\paragraph{Conclusion}

While detectors learned on one dataset may not necessarily transfer
well to others, their ranking is stable across datasets, suggesting
that insights can be learned from well-performing methods regardless
of the benchmark.

\section{\label{sec:Conclusion}Conclusion}

Our experiments show that most of the progress in the last decade
of pedestrian detection can be attributed to the improvement in features
alone. Evidence suggests that this trend will continue. Although some
of these features might be driven by learning, they are mainly hand-crafted
via trial and error. 

Our experiment combining the detector ingredients that our retrospective
analysis found to work well (better features, optical flow, and context)
shows that these ingredients are mostly complementary. Their combination
produces best published detection performance on Caltech-USA.

While the three big families of pedestrian detectors (deformable part
models, decision forests, deep networks) are based on different learning
techniques, their state-of-the-art results are surprisingly close.

The main challenge ahead seems to develop a deeper understanding of
what makes good features good, so as to enable the design of even
better ones.

\bibliographystyle{splncs}
\bibliography{2014_eccvw_ten_years_of_pedestrian_detection}

\chapter*{Ten years of pedestrian detection,\protect \\
what have we learned?\protect \\
Supplementary material}

\section{\label{sec:VJLike-HOGLike-details}Reviewing the effect of features}

The idea behind the experiments in section 4.1 of the main paper is
to demonstrate that, within a single framework, varying the features
can replicate the jump in detection performance over a ten-year span
$(2004-2014)$, i.e. the jump in performance between \texttt{VJ }and
the current state-of-the-art. 
\begin{figure}[t]
\begin{centering}
\hspace*{\fill}\hspace*{-0.5em}\subfloat[\label{fig:effect-of-features-INRIA}INRIA test set]{\begin{centering}
\includegraphics[width=0.5\textwidth]{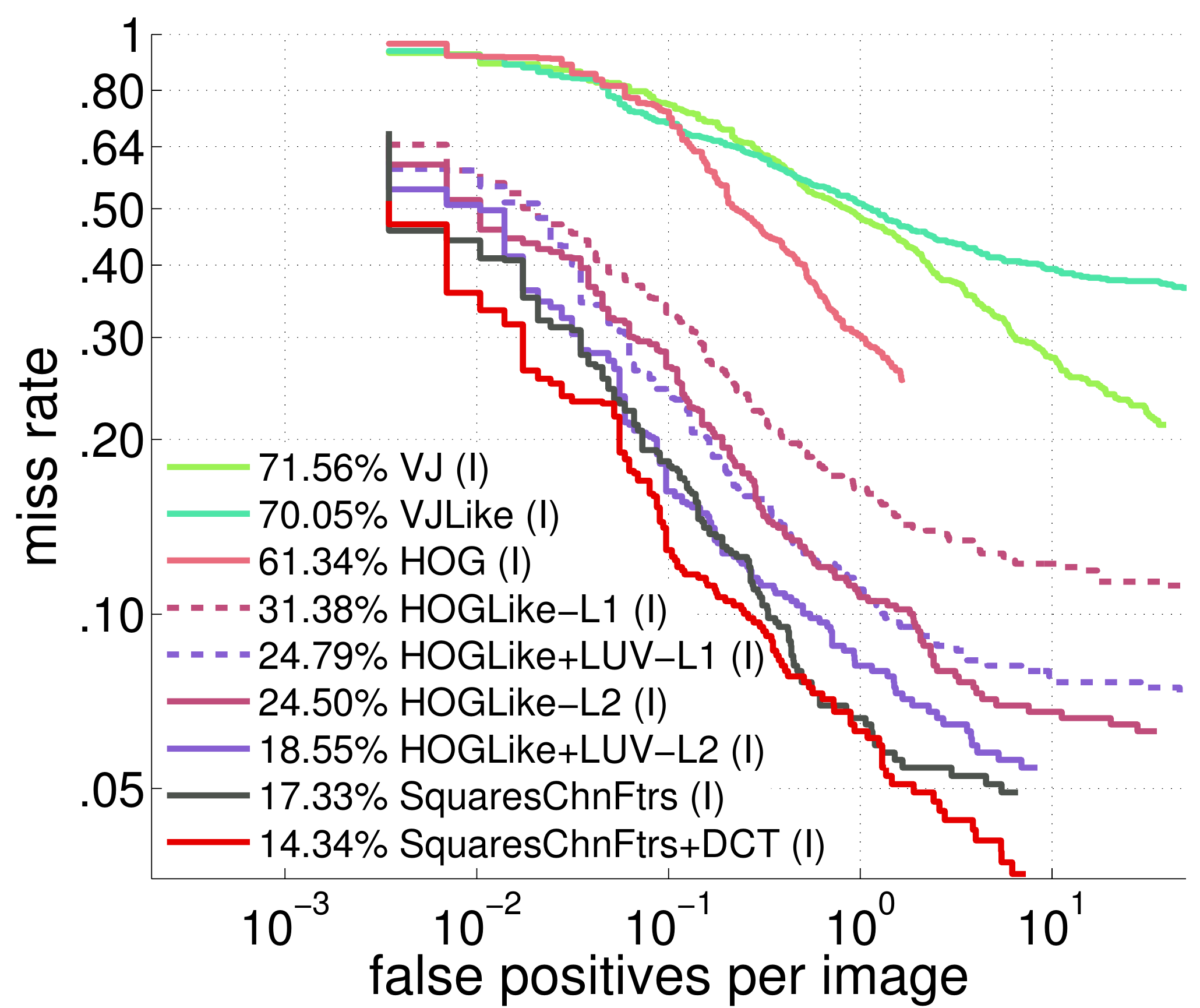}
\par\end{centering}

}\hspace*{\fill}\subfloat[\label{fig:effect-of-features-Caltech-1}Caltech-USA reasonable test
set]{\begin{centering}
\includegraphics[width=0.5\textwidth]{figures/effect_of_features_caltech}
\par\end{centering}

}\hspace*{\fill}
\par\end{centering}

\protect\caption{\label{fig:effect-of-features-inria-and-caltech}Effect of features
on detection performance. (I)/(C) indicates using INRIA/Caltech-USA
training set respectively.}
\end{figure}

See figure \ref{fig:effect-of-features-inria-and-caltech} for results
on INRIA and Caltech-USA of the following methods (all based on \texttt{SquaresChnFtrs},
described in section 4 of the paper):

\begin{lyxlist}{00.00.0000}
\item [{\texttt{VJLike}}] uses only the luminance colour channel, emulating
the original \texttt{VJ} \cite{Viola2003Cvpr}. We use $8\,000$ weak
classifiers to compensate for the weak input feature, only\texttt{
}square pooling regions, and level-2 trees to emulate the Haar wavelet-like
features used by \texttt{VJ}.
\item [{\texttt{HOGLike-L1/L2}}] uses $8\times8\ \mbox{pixel}$ pooling
regions, $6$ oriented gradients, $1$ gradient magnitude, and level
1/2 decision trees (1/3 threshold comparisons respectively). A level-1
tree emulates the non-linearity in the original \texttt{HOG}+linear
SVM features \cite{Dalal2005Cvpr}. 
\item [{\texttt{HOGLike+LUV}}] is identical to \texttt{HOGLike, }but with
additional LUV colour channels ($10$ feature channels total). 
\item [{\texttt{SquaresChnFtrs}}] is the baseline described in the beginning
of the experiments section (\S 4). It is similar to \texttt{HOGLike+LUV}
but the size of the square pooling regions is not restricted.
\item [{\texttt{SquaresChnFtrs+DCT}}] is inspired by \cite{Nam2014arXiv}.
We expand the ten HOG+LUV channels into $40$ channels by convolving
each of the 10 channels with three DCT (discrete cosine transform)
filters ($7\times7\ \mbox{pixels}$), and storing the absolute value
of the filter responses as additional feature channels. The three
DCT basis functions we use as 2d-filters correspond to the lowest
spatial frequencies. We name this variant \texttt{SquaresChnFtrs+DCT}
and it serves as reference point for the performance improvement that
can be obtained by increasing the number of channels.
\end{lyxlist}

\section{\label{sub:Complementarity-experiments-supplementary}Complementarity
of approaches}

Table \ref{tab:Complementarity} contains the detailed results of
combining different approaches with a strong baseline, related to
section 4.2 of the main paper. \texttt{Ka\-ta\-ma\-ri-v1 }combines
all three listed approaches with \texttt{Squa\-res\-Chn\-Ftrs}.
We train and test on the Caltech-USA dataset. It can be noticed that
the obtained improvement is very close to the sum of individual gains,
showing that these approaches are quite complementary amongst each
other.
\begin{table}[h]
\begin{centering}
\protect\caption{\label{tab:Complementarity}Complementarity between different extensions
of the \texttt{Squa\-res\-Chn\-Ftrs} strong baseline. Results in
MR (lower is better). Improvement in MR percent points. Expected improvement
is the direct sum of individual improvements.}

\par\end{centering}

\vspace{0.5em}

\centering{}%
\begin{tabular}{llcccc}
\multirow{2}{*}{Method} & \multirow{2}{*}{\quad{}} & \multirow{2}{*}{Results} & \multirow{2}{*}{\quad{}} & \multirow{2}{*}{Improvement} & Expected \tabularnewline
 &  &  &  &  & improvement\tabularnewline
\hline 
\hline 
\texttt{SquaresChnFtrs} &  & 34.81\% &  & - & -\tabularnewline
\hline 
\texttt{+DCT}  &  & 31.28\% &  & 3.53 & -\tabularnewline
\texttt{+SDt }\cite{Park2010Eccv} &  & 30.34\% &  & 4.47 & -\tabularnewline
\texttt{+2Ped }\cite{Ouyang2013Cvpr} &  & 29.42\% &  & 5.39 & -\tabularnewline
\hline 
\texttt{+DCT+2Ped} &  & 27.40\% &  & 7.41 & 8.92\tabularnewline
\texttt{+SDt+2Ped} &  & 26.68\% &  & 8.13 & 9.86\tabularnewline
\texttt{+DCT+SDt} &  & 25.24\% &  & 9.57 & 8.00\tabularnewline
\hline 
\hline 
\texttt{Katamari-v1} &  & \emph{22.49\%} &  & 12.32 & 13.39\tabularnewline
\end{tabular}
\end{table}

\end{document}